\documentclass{article}
\usepackage{spconf,amsmath,graphicx}
\usepackage{comment}
\usepackage{mathtools}
\usepackage{algorithm2e}
\usepackage{amssymb}
\usepackage{csquotes}
\usepackage{xcolor}
\usepackage{amsmath}
\usepackage{amssymb}
\usepackage{graphicx}
\usepackage{float}
\usepackage{csquotes}
\usepackage{hyperref}
\usepackage{csquotes}
\usepackage{hyperref}
\usepackage{url}
\usepackage{graphicx}
\usepackage{booktabs}
\usepackage{siunitx}

\newcommand*{\ditto}{---\texttt{"}---}
\title{Efficient similarity-based passive filter pruning for compressing CNNs}
\name{Arshdeep Singh, Mark D. Plumbley}%\thanks{Thanks to XYZ agency for funding.}}
\address{Centre for Vision, Speech and Signal Processing (CVSSP)\\
University of Surrey, UK \\
Email: \{arshdeep.singh, m.plumbley\}@surrey.ac.uk}

\begin{document}
\ninept
\RestyleAlgo{ruled}
\SetKwComment{Comment}{/* }{ */}
\maketitle
\begin{abstract}

Convolution neural networks (CNNs) have shown great success in various applications. However, the computational complexity and memory storage of CNNs is a bottleneck for their deployment on resource-constrained devices. Recent efforts towards reducing the computation cost and the memory overhead of CNNs involve similarity-based passive filter pruning methods. Similarity-based passive filter pruning methods compute a  pairwise similarity matrix for the filters and eliminate a few similar filters to obtain a small pruned CNN. However, the computational complexity of computing the pairwise similarity matrix is high, particularly when a convolutional layer has many filters. To reduce the computational complexity in obtaining the pairwise similarity matrix, we propose to use an efficient method where the complete pairwise similarity matrix is approximated from only a few of its columns by using a Nystr{\"o}m approximation method. The proposed efficient similarity-based passive filter pruning method is 3 times faster and gives same accuracy at the same reduction in computations for CNNs compared to that of the similarity-based pruning method that computes a complete pairwise similarity matrix. Apart from this, the proposed efficient similarity-based pruning method performs similarly or better than the existing norm-based pruning methods. The efficacy of the proposed pruning method is evaluated on CNNs such as DCASE 2021 Task 1A baseline network and a VGGish network designed for acoustic scene classification. 

\end{abstract}
\section{Introduction}
\label{sec:intro}

Compressing convolutional neural networks (CNNs) is crucial to reduce their computational complexity and memory storage for efficient deployment on resource-constrained devices \cite{martin2021low}, despite state-of-the-art performances of CNNs in various applications \cite{gu2018recent}. Typically, CNNs have redundant parameters such as weights or filters, which yield only extra computations and storage without contributing much to the performance of the underlying task \cite{denil2013predicting, livni2014computational}. For example, Singh et al. \cite{singh2020svd, singh2019deep} found that  73\% of the filters in SoundNet that do not provide discriminative information across different acoustic scene classes, and eliminating such filters gives similar performance compared to that of using all filters in SoundNet. Thus, the compression of CNNs has recently drawn significant attention from the research community.

Recent efforts towards compressing CNNs involve filter pruning methods \cite{lin2020hrank, luo2018thinet} that eliminate some of the filters in CNNs based on their importance. The importance of the CNN filters is measured in an active or in a  passive manner. Active filter pruning methods involve a dataset.  For example, some methods \cite{luo2017entropy,polyak2015channel,hu2016network} use feature maps which are outputs produced by the filters corresponding to a set of examples, and apply metrics such as entropy or the average percentage of zeros on the feature maps to quantify the filter importance. On the other hand, passive filter pruning methods \cite{li2016pruning, he2019filter} use only parameters of the filters, such as an absolute sum of the weights in the filters, to quantify the filter importance. The passive filter pruning methods do not involve a dataset to measure filter importance and therefore are easier to apply compared to active filter pruning methods. After eliminating filters from the CNNs, the pruned network is fine-tuned to regain some of the performance lost due to the filter elimination.

\begin{figure}[t]
    \centering
    \includegraphics[scale=0.32]{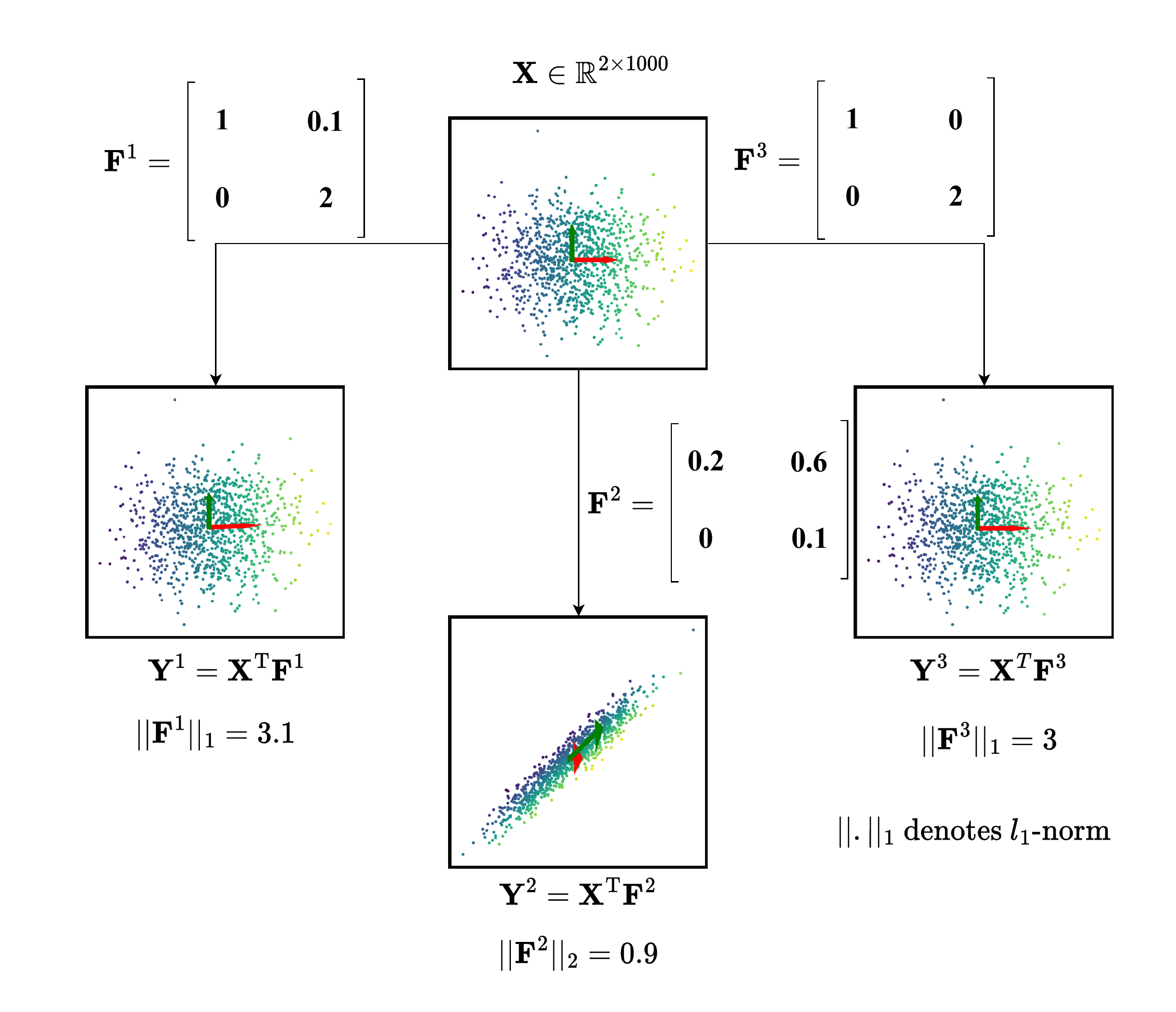}
    \vspace{-0.5cm}
    \caption{An illustration of  output produced in a convolution layer by three CNN filters, $\mathbf{F}^1$, $\mathbf{F}^2$ and $\mathbf{F}^3$,  with a convolution operation on randomly generated data points, $\mathbf{X} \in \mathbb{R}^{2 \times 1000}$.}
    \label{fig: ICASSP illustration}
\end{figure}

Previously, passive filter pruning methods used norm-based metrics such as $l_1$-norm \cite{li2016pruning}, which is a sum of the absolute values of each weight in the filter, or $l_2$-distance  of the filters from a geometric median of all filters \cite{he2019filter} to quantify the importance of the filters. These norm-based methods use a \enquote{smaller-norm-less-important} criterion to eliminate filters. For example, a filter having a relatively high $l_1$-norm is considered more important than others. However, while selecting relatively high-norm filters as important, norm-based methods may ignore the redundancy among the high-norm filters.
%Under situations, where a large number of filters are to be pruned, the norm-based methods always select those filters as important which has relatively high norm which may be redundant. 
To illustrate this, we show outputs produced by three filters in Figure \ref{fig: ICASSP illustration}. Filters $\mathbf{F}^1$ and $\mathbf{F}^3$ have similar $l_1$-norm and produce similar outputs. However, selecting two important filters out of the three filters shown in Figure \ref{fig: ICASSP illustration}, the norm-based method selects  filters $\mathbf{F}^1$ and $\mathbf{F}^3$ as important due to their relatively high norm, despite producing similar outputs, while it eliminates filter $\mathbf{F}^2$ that produces significantly different output than the other filters. Thus the diversity learned in the network may be ignored.

%and eliminate  the filters having low norm which may produce significantly different output compared to that produced by high norm filters as illustrated in Figure \ref{fig: ICASSP illustration}, and hence the diversity learned in the network may be ignored.

%produce different output than that of the high norm filters. An illustration of the significantly varying output produced by a filter with high $l_1$-norm  and a filter with low $l_1$-norm is shown in Figure xxxx.

To capture diversity in the network, similarity-based methods are employed that eliminate similar filters with an assumption that the similar filters produce similar or redundant outputs. For example, Kim et al.  \cite{park2020reprune} perform clustering on filters and selects a filter from each cluster as important and eliminates the other filters. Singh et al. \cite{singh2022passive} measure similarity between filters by computing a pairwise cosine distance for all filters and then eliminating a filter from a pair of similar filters.  Such similarity-based methods give better performance compared to norm-based methods. However, similarity-based pruning methods involve a similarity matrix that takes $\mathcal{O}(n^2d)$ computations to compute for $n$ filters having $d$ parameters. Due to this, the computational complexity is high, particularly when there is a large number of filters in the convolutional layer. 

In this work, we propose passive filter pruning method for CNNs to reduce their computational complexity and memory storage by using a  Nystr{\"o}m approximation \cite{drineas2005nystrom} to approximate the similarity matrix using only a few columns of the complete similarity matrix. 
%Therefore, the computational complexity in computing the pairwise similarity matrix is reduced.
%$O(nmd + m^3)$, which is less.
We evaluate the proposed pruning framework on acoustic scene classification using two CNNs, DCASE 2021 Task 1A baseline network \cite{martin2021low} and VGGish network \cite{hershey2017cnn}.

The rest of this paper is organised as follows. Section \ref{sec: similarity method} explains efficient similarity-based passive filter pruning method. Experimental setup is included in Section \ref{sec: experimental setup}. Section \ref{sec: results} presents results and analysis. Finally, conclusion is included in Section \ref{sec: conclusion}.

%An illustration of output produced by few filters in CNN is shown in Figure . The filter with low-norm  produces significantly different output than that of the other filters with relatively higher norm. Also, two filters with same $l_1$-norm which are consider equally important produces different output. 

%Various feature maps of $\textrm{C3}$ layer of SoundNet; (a)-(f) Showing $i^{th}$ feature map for four audio classes; Shop, hallgare, tubestation, kidegame, (different color represents different class) taken from LITIS Rouen's dataset \cite{rakotomamonjy2015histogram} (a)-(c) illustrate more important feature maps as it shows different responses for different examples. while (d)-(f) illustrate less important feature maps as it shows more or less similar responses for different examples. (B) Distribution of feature maps indexed by 12, 29 and 30 for same audio classes.(g), (h) illustrates the different and (i) shows similar distributions. 

%To measure the similarity among feature maps of different classes, we perform  analysis of variance (ANOVA) test for all feature maps independently. Also differential entropy based framework and cosine similarity based method being performed. In the following section, the proposed methodology is explained to eliminate the similar feature maps.

\section{Efficient Similarity-based passive filter pruning method}
\label{sec: similarity method}

Consider a set of $n$ filters, $\mathbf{F}^l, 1 \le l \le n $ each of size ($w \times h \times c$) with $w$ is a width, $h$ is a height and $c$ is the number of channels, in a convolution layer of a CNN. Each filter is transformed to a  2D matrix of size ($d\times c$) without loss of generality with $d = wh$. Next, we compute a Rank-1 approximation of the filter by performing  singular value decomposition (SVD) on the transformed 2D filter. Next, a column $\in \mathbb{R}^{d}$ with unit norm from the Rank-1 approximation of $\mathbf{F}^l$ is chosen as a representative of the corresponding filter. Let $\mathbf{R} \in \mathbb{R}^{d \times n}$ denotes the filter representative matrix which is constructed by stacking the filter representatives of the  $n$ filters.

%Given $\mathbf{R}$, a distance matrix $\mathbf{Z} \in \mathbb{R}^{n \times n}$  is computed by measuring a cosine distance between each filter pairs in the existing similarity-based pruning method \cite{singh2022passive}.  The computational complexity to obtain $\mathbf{Z}$ is $\mathcal{O}(n^2d)$.  

\begin{algorithm}[ht]
\label{alg: efficiently identification of important filters}
\caption{Efficient similarity-based pruning algorithm to identify important filters in a convolution layer.}
\KwData{Pair-wise similarity matrix of $n$ filters with $m$ filters,  $\mathbf{C}$  $ = [\mathbf{W} \hspace{0.47cm} \mathbf{A}]^\text{T} \in \mathbb{R}^{n\times m}$, $m << n$.} 
\KwResult{Indices of important filters (Imp\_list).}
\textcolor{red}{(Step 1): Obtaining distance matrix via approximating $\mathbf{S}$} 

$\mathbf{W} = \mathbf{U} \Sigma \mathbf{U}^\text{T}$, 

$\mathbf{W}_k^+$  = $\sum_{j=1}^{k}$ $\sigma_j^{-1}  \mathbf{U}^j\mathbf{U}^{j^\text{T}}$, \\
$\tilde{\mathbf{S}} = \mathbf{C}\mathbf{W}_{k}^{+}\mathbf{C}^\text{T}$, 

$\tilde{\mathbf{Z}} = \mathbf{1} - \Tilde{\mathbf{S}}.
$\textcolor{blue}{\hspace{3.2cm}{\%Distance matrix}}

\textcolor{red}{(Step 2): Identify important filter indices} 

Q= [\:], Imp\_list =[\:], Red\_list = [\:] 

\For {$l \leq n$}{
  [$q$, $D$] = argmin\{ $\tilde{\mathbf{Z}}$[ $l$, : - \{$l$\}] \} \textcolor{blue}{\hspace{0.2cm}\%{Identify the closet filter with index $q$ to $l^{\text{th}}$ filter with their distance $D$}} 
  
  Q.append($(l,q)$, $D$) 
}

Q\_sort = Sort(Q)  \textcolor{blue}{{\hspace{0.8cm}\%Sort $\text{Q}$ based on the distance $D$}}

\For {$i \leq len(\text{\textnormal{Q}})$}{
   id\_imp = Q\_sort[$i$][0] \textcolor{blue}{\hspace{1cm}\%{important filter index}} \\
   id\_red = Q\_sort[$i$][1] \textcolor{blue}{\hspace{1cm}\%{redundant filter index}} \\
   \If{\text{\textnormal{id\_imp}} $\notin$ \text{\textnormal{Red\_list}}}{Imp\_list.append( id\_imp )
   
   Red\_list.append( id\_red )
   }
 }%
\end{algorithm}

Given  $\mathbf{R}$, we identify a small set of important filters out of total $n$ filters in a given convolutional layer based on the similarity between the filters using the following two steps:

\noindent \textbf{(Step 1) Approximating distance matrix:} In the first step, we approximate  the pairwise cosine distance matrix $\mathbf{Z} = \mathbf{1} - \mathbf{S}$, where $\mathbf{S} = \mathbf{R}^\text{T}\mathbf{R} \in \mathbb{R}^{n \times n}$ denotes a pairwise similarity matrix for $n$ filters.  %$\mathbf{S}$  is a symmetric positive semidefinite (SPSD) matrix as well.  

We take a few columns of $\mathbf{S}$ to approximate the rest of the entries of $\mathbf{S}$ by using a Nystr{\"o}m approximation method \cite{drineas2005nystrom}. 
Without loss of generality, the matrix $\mathbf{S}$ can be written as follows:

\begin{equation}
\centering
\mathbf{S} = \begin{bmatrix}
 \mathbf{W} & \mathbf{A}^{\text{T}} \\
 \mathbf{A} & \mathbf{B}  \\
\end{bmatrix} \textrm{and   }  \mathbf{C} = \begin{bmatrix}
\mathbf{W} \\
\mathbf{A}
\end{bmatrix} 
\end{equation}

\noindent where $\mathbf{W} \in \mathbb{R}^{m \times m}$,   $\mathbf{C} \in \mathbb{R}^{n \times m}$, $\mathbf{A} \in \mathbb{R}^{(n-m)\times m}$, and $m << n$.  

The Nystr{\"o}m method  approximates $\mathbf{S}$  by taking  $\mathbf{C}$, $m$ columns  from $\mathbf{S}$,  generating a rank-$k$ approximation $\Tilde{\mathbf{S}}$ of $\mathbf{S}$ given by,

\begin{equation}
    \Tilde{\mathbf{S}} = \mathbf{C}\mathbf{W}_{k}^{+}\mathbf{C}^\text{T},
\end{equation}

\noindent where $\mathbf{W}_k$ is the best rank-$k$ approximation of $\mathbf{W}$ for the Frobenius norm with $k \le \text{rank}(\mathbf{W})$ and  $\mathbf{W}_k^+$  = $\sum_{j=1}^{k}$ $\sigma_j^{-1}  \mathbf{U}^j\mathbf{U}^{j^\text{T}}$ denotes the pseudo-inverse of $\mathbf{W}_k$. $\mathbf{W}_k^+$ is obtained  by performing SVD on $\mathbf{W} = \mathbf{U} \Sigma \mathbf{U}^\text{T}$, where $\mathbf{U}$ is an orthonormal matrix,  $\mathbf{U}^j$ is an $j^{\text{th}}$ column of $\mathbf{U}$ and $\Sigma = \text{diag}\{\sigma_1, \sigma_2,\dots,\sigma_m\}$ is a real diagonal matrix with $\sigma_1 \ge \sigma_2, \dots,  \sigma_m \ge 0$. The computational complexity needed to obtain $\Tilde{\mathbf{S}}$  is $\mathcal{O}(m^3 + nmk)$.  After obtaining $\Tilde{\mathbf{S}}$, we compute $\tilde{\mathbf{Z}} = \mathbf{1} - \tilde{\mathbf{S}}$, as an approximation of $\mathbf{Z}$.

\noindent \textbf{(Step 2) Obtaining important filters:} Given $\tilde{\mathbf{Z}}$, we identify the closet filter corresponding to each filter. A filter from the closest filter pairs is then considered redundant and eliminated from the  underlying convolution layer. 

A summary of the overall framework  is given in Algorithm \ref{alg: efficiently identification of important filters}.

\noindent \textbf{Obtaining pruned network and performing fine-tuning:} 
After obtaining  the important filters across different convolution layers using Algorithm \ref{alg: efficiently identification of important filters}, we retain the set of important filters and eliminate the  other filters from the unpruned CNN to obtain a pruned network.  Eliminating a filter from a given convolutional layer also removes the corresponding feature map produced by the filter and the associated channel of the filter in the following convolutional layer. Therefore, the computations in the next convolutional layer are also reduced in the pruned network.

After removing filters, we perform fine-tuning which involves re-training of the pruned network to regain some of the lost performance due to the removal of the connection from the unpruned CNN. The codes for the proposed efficient pruning framework can be found at the link\footnote{\url{https://github.com/Arshdeep-Singh-Boparai/Efficient_similarity_Pruning_Algo.git}}.

\section{Experimental setup}
\label{sec: experimental setup}

We evaluate the proposed pruning framework on CNNs designed for acoustic scene classification (ASC). An overview of the unpruned CNNs is given below,

\begin{figure}[t]
    \centering
    \includegraphics[scale=0.52]{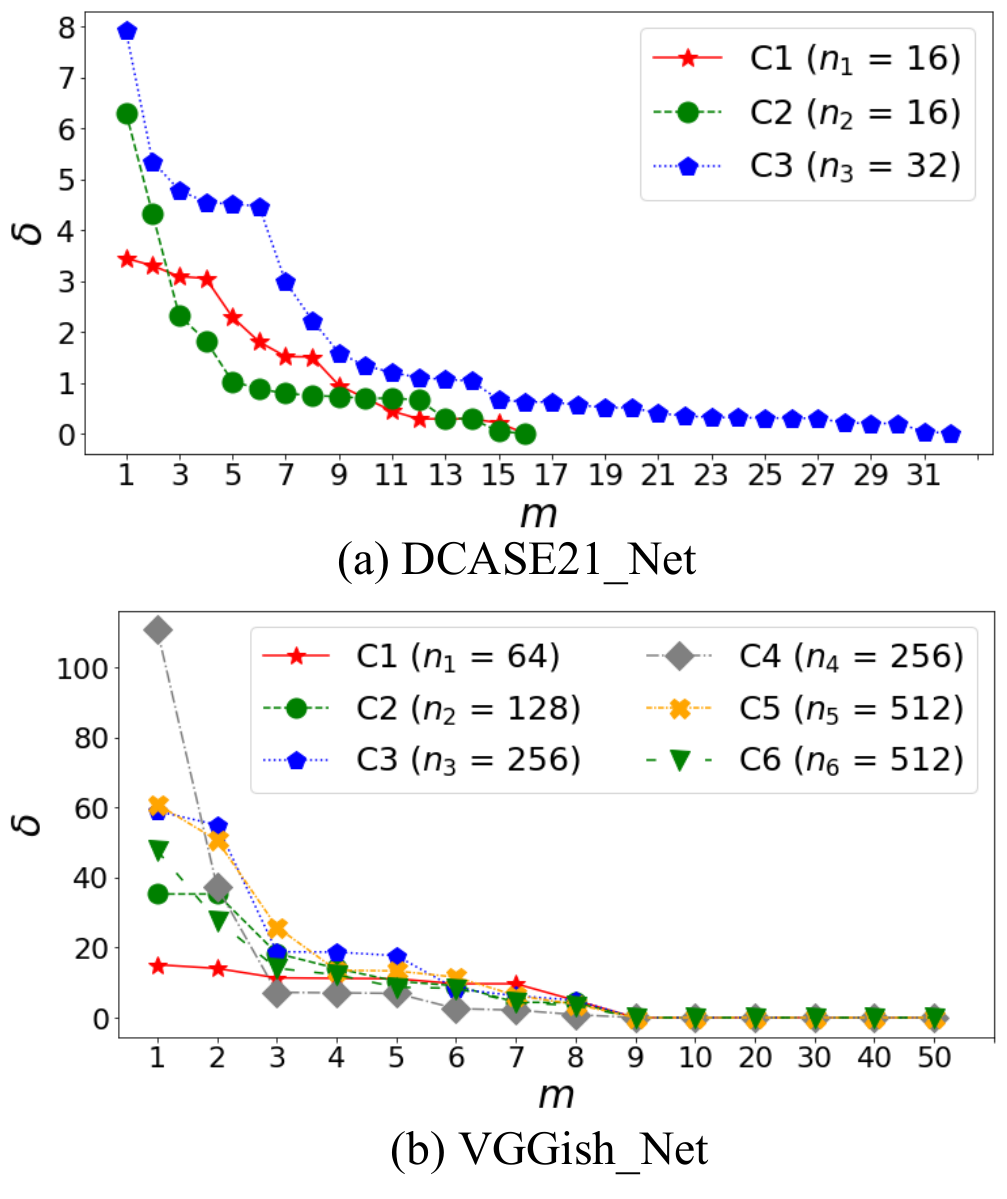}
    \vspace{-0.5cm}
    \caption{Approximation error ($\delta$) when $m$ columns are selected out of $n$ columns from the similarity matrix for different convolutional layers in (a) DCASE21\_Net and (b) VGGish\_Net . Here, the similarity matrix is computed using rank-$k$ approximation with  $k = m$.}
    \label{fig: episoln vs m...}
\end{figure}

\textbf{(a) DCASE21\_Net:} DCASE21\_Net is a publicly available pre-trained network designed for DCASE 2021 Task 1A that is trained using TAU Urban Acoustic Scenes 2020 Mobile development dataset (we denote \enquote{DCASE-20}) to classify 10 different acoustic scenes \cite{martin2021low}. The input to the network is a log-mel spectrogram of size (40 × 500) corresponding to a 10s audio clip. DCASE21\_Net is trained using the Adam optimizer with cross-entropy loss function for 200 epochs. The trained network  has 46,246 parameters and requires approximately 287M multiply-accumulate operations (MACs) during inference corresponding to 10-second-length audio clip, and gives 48.58\% accuracy on the DCASE-20 development validation dataset. DCASE21\_Net consists of three convolutional layers (termed as C1 to C3) and one fully connected layer.  C1  has $n_1 =16$, C2 has $n_2 =16$ and C3 has $n_3 = 32$ filters. 

%The number of filters in each convolutional layers are 16,16,32 respectively.

\textbf{(b) VGGish\_Net:} This is built using a publicly available pre-trained VGGish network \cite{hershey2017cnn} followed by a dense and a classification layer. We train VGGish\_Net on the TUT Urban Acoustic Scenes 2018 development (\enquote{DCASE-18}) training dataset \cite{Mesaros2018_DCASE} to classify 10 different acoustic scenes using Adam optimizer with cross-entropy loss function for 200 epochs. The input to the VGGish\_Net is a log-mel spectrogram of size (96 $\times$ 64) computed corresponding to a 960ms audio segment from a whole 10s audio scene. The VGGish\_Net has approximately 55.361M parameters and requires 903M MACs during inference corresponding to an audio clip of 960ms and gives  64.69\% accuracy on 10s audio scene for DCASE-18 development validation dataset. VGGish\_Net has six convolution layers (termed as C1 to C6). The number of filters in each convolutional layers are \{64, 128, 256, 256, 512, 512\} respectively.

For $i^{\text{th}}$ convolutional layer, we approximate the distance matrix $\tilde{\mathbf{Z}}_{m_i,k_i}$  using  first 1 to $m_i$ columns of the similarity matrix $\mathbf{S}_i$ and approximating the similarity matrix by rank-$k_i$ approximation where 1 $\le k_i \le m_i$. To measure the effectiveness of the approximation, we compute an approximation error $\delta_i = ||\mathbf{Z}_i - \tilde{\mathbf{Z}}_{m_i,k_i}||_2$ at different values of $m_i$ and $k_i$.

%We also analyse $\delta_i$ as a function of $k_i$ by varying $k_i$ from 1 to $m_i$ for $i^{\text{th}}$ convolutional layer.

%to analyse an optimal value of $m$, where the important set of filters obtained by involving (Step 1 + Step 2) in the Algorithm 1 are equal to the important set of the filters obtained using the original distance matrix ($\mathbf{Z}$) computed without any approximation and using  Step 2 in Algorithm \ref{alg: efficiently identification of important filters}.

%We obtain a pruned network by retaining an important set of filters obtained using Algorithm \ref{alg: efficiently identification of important filters} and eliminating the other filters from the unpruned network. 
To obtain the pruned network, we identify a set of important filters by computing $\tilde{\mathbf{Z}}_{m_i,k_i}$ at $m_i$ and $k_i$, where $\delta_i$ $<$ 1. Fine-tuning of the pruned network is performed  with similar conditions such as loss function, optimizer as used for training the unpruned network except for 100 epochs.

\noindent \textbf{Performance metrics:} We analyse a total  time required to obtain the set of important filters for all convolutional layer. The total pruning time is  computed after running the pruning algorithm for 10K times and an average of the total pruning time is reported. To measure the performance of the pruned network, we compute accuracy, the number of MACs per inference and the number of parameters. The accuracy of the pruned network is computed after fine-tuning the pruned network independently for 5 times and we report the average accuracy.

\noindent \textbf{Other methods for comparison}: 
We compare the  proposed Algorithm \ref{alg: efficiently identification of important filters} with existing norm-based pruning methods such as an $l_1$-norm \cite{li2016pruning} method and a geometric median (GM) method \cite{he2019filter}, and a similarity-based pruning method \cite{singh2022passive} that first computes complete cosine distance matrix, and then uses Step 2 of the Algorithm \ref{alg: efficiently identification of important filters} to compute important set of filters for a given convolutional layer.

%We compare the accuracy of the pruned network obtained using the proposed Algorithm \ref{alg: efficiently identification of important filters} with that of the similar size pruned network obtained using the existing norm-based pruning methods such as an $l_1$-norm \cite{li2016pruning} method and a geometric median (GM) method \cite{he2019filter}, and a similarity-based pruning method \cite{singh2022passive} that uses original cosine distance matrix ($\mathbf{Z}$) and Step 2 from the Algorithm \ref{alg: efficiently identification of important filters} to compute important set of filters.  

\begin{figure}[t]
    \centering
    \includegraphics[scale=0.5]{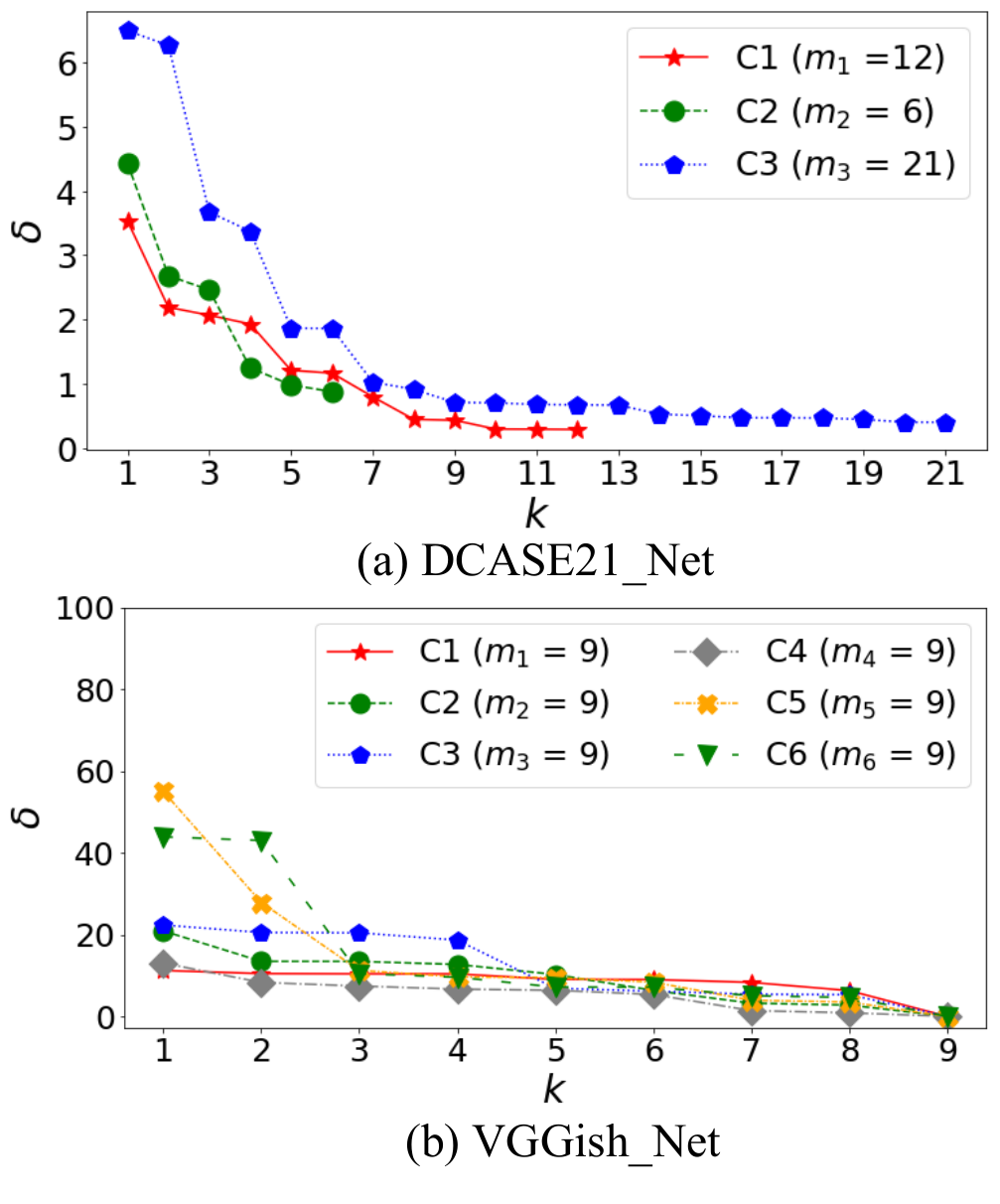}
    \vspace{-0.5cm}
    \caption{Approximation error ($\delta$) when the similarity matrix is generated with rank-$k$ approximation by varying $k$ using fixed number of columns ($m$) across various convolutional layers for (a) DCASE21\_Net having $m_1 =$12 for C1, $m_2 =$ 6 for C2 and $m_3 =$21 for C3 and (b) VGGish\_Net where $m$ = 9 for all C1 to C6 layers.}
    \label{fig: episoln vs k}
\end{figure}

\section{Results and Analysis}
\label{sec: results}

\begin{table*}[ht]
	\caption{Comparison of accuracy,  MACs and parameters  among  the (a) unpruned, (b) the pruned networks obtained using various pruning methods. Total pruning time to obtain important set of filters across all convolutional layers is also shown.}
	\centering
	\resizebox{0.78\textwidth}{!}{
		\begin{tabular}{SSSSSS} \toprule
			{Network} & {Pruning Method} & {Accuracy (\%)}  & {MACs} &  {Parameters} & {Total pruning time (s)} \\ \midrule
   			{DCASE21\_Net} & {No pruning} &{48.58} & {287M} & {46246} & {-}  \\

                     {on DCASE-20 dataset} & {$l_1$-norm \cite{li2016pruning}}  & {44.42} & {139M}  & {24056} & {0.0072} \\
			          {} &  {GM \cite{he2019filter}} & {45.84} & {\ditto} & {\ditto} & {0.010} \\
                      {} & {Similarity-based \cite{singh2022passive}} & {45.54} & {\ditto} & {\ditto} & {0.063} \\
                      {} & { (Proposed) Efficient similarity-based} & {45.54} & {\ditto} & {\ditto} & {0.011} \\
             \midrule
			{VGGish\_Net} & {No pruning} &{64.69} & {903M} & {55M} & {-}  \\

                  {on DCASE-18 dataset} & {$l_1$-norm \cite{li2016pruning}}  & {60.02} & {595M}  & {42.89M} & {0.21} \\
			      {} &  {GM \cite{he2019filter}} & {59.71} & {\ditto} & {\ditto} & {0.42} \\
			      {} & {Similarity-based \cite{singh2022passive}} & {62.00} & {\ditto} & {\ditto} & {34.80} \\
			      {} & {(Proposed) Efficient similarity-based} & {62.00} & {\ditto} & {\ditto} & {11.70} \\

   \bottomrule
 	\end{tabular}}
	\label{tab: VGGish  COMPARISON L1 GM PCS}
\end{table*}

\begin{figure}[t]
    \centering
    \includegraphics[scale=0.56]{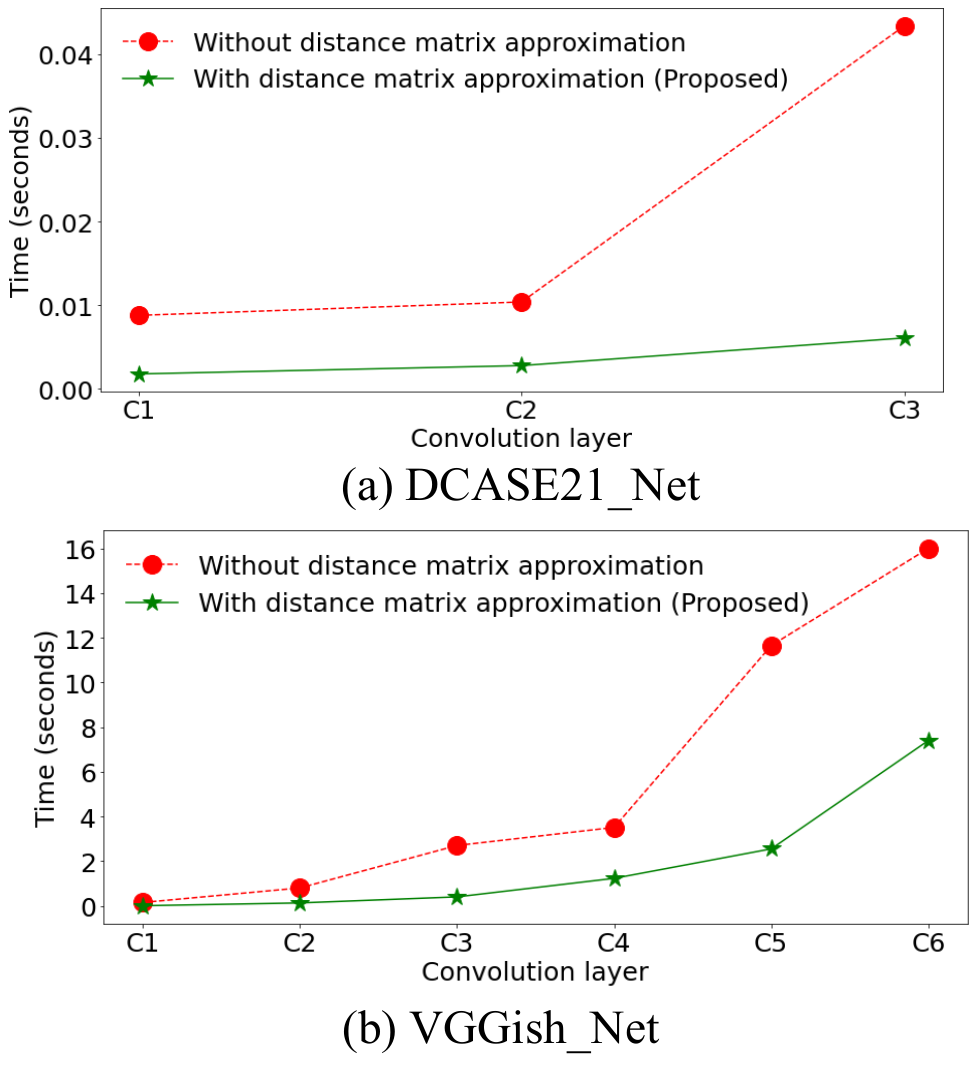}  
        \vspace{-0.5cm}
    \caption{Total pruning time to obtain important set of filters using Algorithm \ref{alg: efficiently identification of important filters} for various convolutional layers, with and without approximating the distance matrix.}
    \label{fig: computational time}
\end{figure}

Figure \ref{fig: episoln vs m...} shows the approximation error when $m_i$ columns  are selected out of $n_i$ columns from the similarity matrix  and the similarity matrix is approximated using rank-$k_i$ approximation with $k_i = m_i$ for different convolutional layers in
DCASE21\_Net and VGGish\_Net.  We observe that selecting few columns from the similarity matrix are sufficient to approximate the complete similarity matrix with $\delta_i < 1$. Also, the distance matrix  ($\tilde{\mathbf{Z}}_{m_i,m_i}$) approximated by choosing $m_i$ columns of the similarity matrix that gives $\delta_i < 1$ results in a same set of important filters  as obtained using the complete distance matrix ($\mathbf{Z}_i$).

For DCASE21\_Net as shown in Figure \ref{fig: episoln vs m...}(a), we find that choosing $m_1 = $ 12 out of $n_1 =$ 16 columns for C1, $m_2 =$ 6 out of $n_2 =$ 16 columns for C2 and $m_3 =$ 21 out of $n_3 =$ 32 columns for C3 gives $\tilde{\mathbf{Z}}_{m_i,m_i} \approx \mathbf{Z}_i$. For VGGish\_Net as shown in Figure \ref{fig: episoln vs m...}(b), we find that  choosing $m_i =$  9 out of $n_i$ columns, where $n_i \in $ \{64, 128, 256, 256, 512, 512\} for C1 to C6 layers respectively, gives $\tilde{\mathbf{Z}}_{m_i,m_i} \approx \mathbf{Z}_i$.

Next, Figure \ref{fig: episoln vs k}  shows the approximation error as $k_i$ varies from 1  to $m_i$ across different convolutional layers of (a) DCASE21\_Net, where $m_1 =$  12, $m_2 =$  6 and $m_3 =$ 21 for C1, C2 and C3 layers respectively, and (b) VGGish\_Net, where $m_i =$ 9 across various convolutional layers. We obtain the same set of important filters using $\tilde{\mathbf{Z}}_{m_i,k_i}$ as that of $\mathbf{Z}_i$  with $\delta_i < 1$ when %We obtain same important filters by using $\tilde{\mathbf{Z}}$ as that of  $\mathbf{Z}$, when
 $k_1 =$ 9 for C1, $k_2 =$  6  for C2 and $k_3 =$  13 for C3 layer in DCASE21\_Net, and $k_i =$ 9 for $1 \le i \le 6$ convolutional layers of VGGish\_Net.

Figure \ref{fig: computational time} compares the total pruning time computed for each convolutional layer, when the distance matrix ($\tilde{\mathbf{Z}}_{m_i,k_i} \text{with } \delta_i < 1$) is approximated using Algorithm \ref{alg: efficiently identification of important filters}: Step 1, and when the complete pairwise distance matrix  ($\mathbf{Z}_i$) is computed without any approximation for (a) DCASE21\_Net and (b) VGGish\_Net.

The total pruning time is reduced by approximating the distance matrix compared to computing the complete pairwise distance matrix for various convolution layers. When the number of filters is large, for example, the C6 layer in VGGish\_Net has 512 filters, the total pruning time  reduces significantly with the distance matrix approximation to that of  computing the complete distance matrix. On the other hand, when the number of filters is smaller, for example C1 layer of VGGish\_Net has 64 filters or all convolutional layers in DCASE21\_Net has  $n_i \le $ 32, the total pruning time reduces marginally by approximating the distance matrix compared to that of computing the complete distance matrix. 

%For convolutional layers with large number of filters, for example C6 layer in VGGish has 512 filters, the total pruning time with similarity matrix approximation is significantly smaller than that of without any approximation 

Table \ref{tab: VGGish  COMPARISON L1 GM PCS} compares the performance metrics with the other methods. For DCASE21\_Net, the pruned network obtained using the proposed pruning method reduces both the MACs and the parameters by approximately 50\% at 3 percentage points drop in accuracy compared to the unpruned network. The total pruning time for $l_1$-norm method \cite{li2016pruning} is the smallest among other methods. However, the accuracy obtained using the $l_1$-norm method is 1 percentage points lesser than that of the proposed pruning method. The accuracy and the total pruning time for the geometrical median (GM) pruning method \cite{he2019filter} is marginally better than that of the proposed pruning method. In contrast to the similarity-based pruning method \cite{singh2022passive}, the proposed efficient similarity-based pruning method takes less total pruning time and gives similar accuracy.

For VGGish\_Net, the pruned network obtained using the proposed pruning method reduces the MACs by 34\%, and the parameters are reduced by 23\% at 2.7 percentage points drop in the accuracy compared to that of the unpruned network. Even though the  $l_1$-norm and the GM pruning methods take significantly less computations than the proposed pruning method, the proposed pruning method improves the accuracy of the pruned network by 2 percentage points compared to that of the  $l_1$-norm and the GM pruning methods. In contrast to the similarity-based pruning method, the proposed efficient similarity-based pruning method is  3 times faster and gives the same accuracy.

\section{Conclusion}
\label{sec: conclusion}

This paper presents an efficient similarity-based passive filter pruning framework to reduce computational complexity and memory storage in CNNs. We show that using only a few columns of the similarity matrix is sufficient to approximate similarity matrix and is 3 times faster than  computing the complete pairwise similarity matrix with no loss in accuracy. The proposed pruning method yields a pruned network that performs similarly or better than the existing norm-based pruning methods.

In future, we would like to improve the performance of the pruned network obtained using the proposed pruning framework to achieve a similar performance as that of the unpruned network by using better distance measures such as graph-based similarity between the filters. Also, reducing the number of fine-tuning epochs (e.g. $ <100$) to recover some of the  performance lost due to filter elimination is a future goal to reduce overall computations.

\section{Acknowledgements}

This work was partly supported by Engineering and Physical Sciences Research Council (EPSRC) Grant EP/T019751/1 \enquote{AI for Sound (AI4S)}. For the purpose of open access, the authors have applied a Creative Commons Attribution (CC BY) licence to any Author Accepted Manuscript version arising.

\newpage
\bibliographystyle{IEEEbib}
\bibliography{ref1.bib}

\begin{thebibliography}{10}

\bibitem{martin2021low}
Irene Mart{\'\i}n-Morat{\'o}, Toni Heittola, Annamaria Mesaros, and Tuomas
  Virtanen,
\newblock ``{Low-complexity acoustic scene classification for multi-device
  audio: Analysis of DCASE 2021 Challenge systems},''
\newblock {\em DCASE Workshop}, 2021.

\bibitem{gu2018recent}
Jiuxiang Gu, Zhenhua Wang, Jason Kuen, Lianyang Ma, Amir Shahroudy, Bing Shuai,
  Ting Liu, Xingxing Wang, Gang Wang, Jianfei Cai, et~al.,
\newblock ``{Recent advances in convolutional neural networks},''
\newblock {\em Pattern Recognition}, vol. 77, pp. 354--377, 2018.

\bibitem{denil2013predicting}
Misha Denil, Babak Shakibi, Laurent Dinh, M.A. Ranzato, and Nando De~Freitas,
\newblock ``Predicting parameters in deep learning,''
\newblock {\em Advances in Neural Information Processing Systems}, pp.
  2148--2156, 2013.

\bibitem{livni2014computational}
Roi Livni, Shai Shalev-Shwartz, and Ohad Shamir,
\newblock ``On the computational efficiency of training neural networks,''
\newblock {\em Advances in Neural Information Processing Systems}, pp.
  855--863, 2014.

\bibitem{singh2020svd}
Arshdeep Singh, Padmanabhan Rajan, and Arnav Bhavsar,
\newblock ``{S{VD}-based redundancy removal in 1-D CNNs for acoustic scene
  classification},''
\newblock {\em Pattern Recognition Letters}, vol. 131, pp. 383--389, 2020.

\bibitem{singh2019deep}
Arshdeep Singh, Padmanabhan Rajan, and Arnav Bhavsar,
\newblock ``Deep hidden analysis: A statistical framework to prune feature
  maps,''
\newblock {\em IEEE International Conference on Acoustics, Speech and Signal
  Processing (ICASSP)}, pp. 820--824, 2019.

\bibitem{lin2020hrank}
Mingbao Lin, Rongrong Ji, Yan Wang, Yichen Zhang, Baochang Zhang, Yonghong
  Tian, and Ling Shao,
\newblock ``{HRank: Filter pruning using high-rank feature map},''
\newblock {\em Proceedings of the IEEE/CVF conference on Computer Vision and
  Pattern Recognition}, pp. 1529--1538, 2020.

\bibitem{luo2018thinet}
Jian-Hao Luo, Hao Zhang, Hong-Yu Zhou, Chen-Wei Xie, Jianxin Wu, and Weiyao
  Lin,
\newblock ``{ThiNet: Pruning {CNN} filters for a thinner net},''
\newblock {\em IEEE Transactions on Pattern Analysis and Machine Intelligence},
  vol. 41, no. 10, pp. 2525--2538, 2018.

\bibitem{luo2017entropy}
Jian-Hao Luo and Jianxin Wu,
\newblock ``An entropy-based pruning method for {CNN} compression,''
\newblock {\em arXiv preprint arXiv:1706.05791}, 2017.

\bibitem{polyak2015channel}
Adam Polyak and Lior Wolf,
\newblock ``Channel-level acceleration of deep face representations,''
\newblock {\em IEEE Access}, vol. 3, pp. 2163--2175, 2015.

\bibitem{hu2016network}
Hengyuan Hu, Rui Peng, Yu-Wing Tai, and Chi-Keung Tang,
\newblock ``Network trimming: A data-driven neuron pruning approach towards
  efficient deep architectures,''
\newblock {\em arXiv preprint arXiv:1607.03250}, 2016.

\bibitem{li2016pruning}
Hao Li, Asim Kadav, Igor Durdanovic, Hanan Samet, and Hans~Peter Graf,
\newblock ``{Pruning filters for efficient ConvNets},''
\newblock {\em International Conference on Learning Representations}, 2017.

\bibitem{he2019filter}
Yang He, Ping Liu, Ziwei Wang, Zhilan Hu, and Yi~Yang,
\newblock ``Filter pruning via geometric median for deep convolutional neural
  networks acceleration,''
\newblock {\em Proceedings of the IEEE Conference on Computer Vision and
  Pattern Recognition}, pp. 4340--4349, 2019.

\bibitem{park2020reprune}
Mincheol Park, Woojeong Kim, and Suhyun Kim,
\newblock ``R{EP}rune: Filter pruning via representative election,''
\newblock {\em arXiv preprint arXiv:2007.06932}, 2020.

\bibitem{singh2022passive}
Arshdeep Singh and Mark~D Plumbley,
\newblock ``A passive similarity based {CNN} filter pruning for efficient
  acoustic scene classification,''
\newblock {\em Interspeech (arXiv preprint arXiv:2203.15751)}, 2022.

\bibitem{drineas2005nystrom}
Petros Drineas, Michael~W Mahoney, and Nello Cristianini,
\newblock ``{On the Nystr{\"o}m method for approximating a Gram matrix for
  improved kernel-based learning},''
\newblock {\em Journal of Machine Learning Research}, vol. 6, no. 12, 2005.

\bibitem{hershey2017cnn}
Shawn Hershey, Sourish Chaudhuri, Daniel~PW Ellis, Jort~F Gemmeke, Aren Jansen,
  R~Channing Moore, Manoj Plakal, Devin Platt, Rif~A Saurous, Bryan Seybold,
  et~al.,
\newblock ``{CNN architectures for large-scale audio classification},''
\newblock {\em International Conference on Acoustics, Speech and Signal
  Processing (ICASSP)}, pp. 131--135, 2017.

\bibitem{Mesaros2018_DCASE}
Annamaria Mesaros, Toni Heittola, and Tuomas Virtanen,
\newblock ``A multi-device dataset for urban acoustic scene classification,''
\newblock {\em Proceedings of the Detection and Classification of Acoustic
  Scenes and Events 2018 Workshop}, pp. 9--13, November 2018.

\end{thebibliography}

\end{document}